\documentclass[letterpaper]{article}
\usepackage{aaai}
\usepackage{amsmath}
\usepackage{times}
\usepackage{helvet}
\usepackage{courier}
\usepackage{todonotes}
\usepackage{multirow}	
\usepackage{subcaption}
\usepackage[ruled,vlined,linesnumbered]{algorithm2e}
\begin{document}
%
\title{SEER: Supervised Learning to Control Energetic Reasoning\thanks{This paper was presented at the CP2014 Doctoral Program of Lyon.}}
\author{Sascha Van Cauwelaert$^1$, Michele Lombardi$^2$, and Pierre Schaus$^1$\\
Universit\'e Catholique de Louvain,
Universit\`a di Bologna}
\maketitle
\begin{abstract}
\begin{quote}
One of the main strengths of Constraint Programming is the ability to reduce the search space via propagation. However, propagation is a double-edged sword, with more pruning power coming at the price of larger computation time. For each problem constraint, the best propagator depends on the specific instance and may change at search time. In the literature, Machine Learning (ML) techniques and activity-based heuristics have been applied respectively for choosing (statically) the propagators for a batch of problems and to adapt (dynamically) the propagation strength. We propose to merge those efforts by using an oracle function, obtained via ML, to decide whether to run complex propagators for a target constraint. A combination of design choices makes the approach flexible and easy to embed in state-of-the-art solvers. In this paper, we focus on investigating the feasibility of building an oracle for the Energetic Reasoning propagator. Our experiments show that high prediction accuracy can be obtained, provide suggestions for classification features, and highlight important issues to address when building such an oracle.
\end{quote}
\end{abstract}


\newcommand{\est}{\mathit{est}}
\newcommand{\lst}{\mathit{lst}}
\newcommand{\ect}{\mathit{ect}}
\newcommand{\lct}{\mathit{lct}}
\newcommand{\dur}{\mathit{dur}}

\newcommand{\mys}{\boldmath{s}}
\newcommand{\myd}{\boldmath{d}}
\newcommand{\mye}{\boldmath{e}}

\newcommand{\mysp}{\boldmath{s'}}
\newcommand{\mydp}{\boldmath{d'}}
\newcommand{\myep}{\boldmath{e'}}

\newcommand{\myspp}{\boldmath{s''}}
\newcommand{\mydpp}{\boldmath{d''}}
\newcommand{\myepp}{\boldmath{e''}}

\newcommand{\classifier}{\mathit{classifier}}	

\newcommand{\TP}{\mathit{TP}}
\newcommand{\TN}{\mathit{TN}}
\newcommand{\FP}{\mathit{FP}}
\newcommand{\FN}{\mathit{FN}}



\section{Introduction} 
\label{sec:introduction}


The ability to reduce the size of the search space via propagation is one of the distinguishing features of Constraint Programming (CP), accounting for much of its ability to solve complex combinatorial problems. Propagation, however, is a double-edged sword: more powerful filtering algorithms provide an increased chance to prune values, but they also have larger computation time, that must be paid regardless of whether additional propagation is actually achieved.

For instance, the \texttt{cumulative} constraint is widely employed to model resource restrictions for scheduling problems with non-preemptive activities. The constraint is one of the best studied in CP and counts a wide range of propagators. In particular, Timetable-based algorithms (such as SWEEP \cite{beldiceanu2002new}) represent the lightweight end of the spectrum, with a time complexity of $O(n^2)$. Timetable propagation (TT) is known to be dominated by Energetic Reasoning (ER, see \cite{baptiste2001constraint}), which is however seldom used in practice because the algorithm has higher time complexity -- $O(n^3)$ -- and often ends up with the same domain reduction as the Timetable.

Nevertheless, there are instances where ER can dramatically reduce the search tree. On the BL benchmarks \cite{baptiste2000constraint}, such propagator may reduce the number of backtracks by a factor of 10 on $\sim 65\%$ of the instances when using dichotomic search on the start variables, and on $\sim 88\%$ when using scheduling-specific search -- \emph{SetTimes}, see \cite{lepape1994settimes}. A similar behavior occurs whenever a constraint has propagators with different pruning power and complexity.


The best propagator depends on specificities of the target problem and instance, and it is far from trivial to select. The choice is typically done by the model designer based on personal experience, intuition, and pilot tests, with mixed outcomes  (we refer to \cite{smith2006modelling} for interested reader). Recently, Machine Learning (ML) techniques have been proposed as a way to automatize the decision \cite{gent2010machine}, with promising results. Such an approach, however, does not take into account the impact of search decisions on the propagator effectiveness. This has been recognized in \cite{stergiou2009heuristics}, where the authors propose to adjust the propagation strength at run time via heuristics based on the solver state. Unfortunately, designing and choosing the correct heuristics is a complex (and problem-dependent) task. Parametrized consistency is introduced in \cite{balafrej2013adaptive}, where a local consistency property is compared to a threshold parameter in order to enforce different consistency levels on values. Heuristics are considered to dynamically adapt the parameter.


\emph{We propose to merge the mentioned approaches and use an oracle, obtained via ML, to predict at run time if running a specific propagator for a constraint will be beneficial. The decision should be based on the current domain of the variables in the constraint scope, i.e. on the input of the propagator itself}. For example such an oracle (in fact, a classifier) could be used to decide for a cumulative constraint whether or not to run ER after a TT propagator has reached the fix point. Compared to \cite{gent2010machine}, our approach can be used to adjust the propagation level at search time. Compared to \cite{stergiou2009heuristics}, it is less-problem dependent, because it operates at a single-constraint level. Moreover, the use of ML spares the designer the effort to find a good heuristic rule (though feature selection is still a problem).

Working at a single constraint level makes our approach simple to implement on state-of-the-art solvers. Additionally, since the prediction we consider concerns a single propagator, there is no need to retrain it when a new propagation algorithm is introduced. Finally, since the oracle input does not include state information (except of course for the domains), the method is well suited for complex search techniques such as non-chronological backtracking and (more importantly) Large Neighborhood Search.

Deploying effectively such an approach is an ambitious endeavor. As a start, in this paper we investigate the feasibility of building an oracle for the ER propagator, for which we use the acronym SEER (Supervised lEarning to control Energetic Reasoning). In particular, we focus on the problem of detecting whether running ER will narrow the domains after TT has reached a fix-point. Temporal aspects (e.g. the trade-off between the amount of propagation and the reduction of the solution time) are left for future research. In this study, we show that high prediction rates can indeed be obtained, we provide guidelines about how to define good training sets, we suggest effective features to be used as input for the classifier, and we highlight critical issues to be addressed in the design of the oracle.

\section{Background and Related Work} 
\label{sec:background_and_related_work}

\subsection{CP and Scheduling Problems} 
\label{sub:cp_and_scheduling_problems}

Constraint Programming is a technique to solve Constraint Satisfaction Problems (CSP). A CSP is a triple $\langle X, D, C\rangle$, where $X$ is a set of variables $x_i$, $D$ is the set of their domains $D_i$ (typically finite and integer), and $C$ is a set of constraints $c_k$ that must be satisfied. Each constraint is defined over a subset of variables $S(c_k)$, called scope, and has an associated algorithm (propagator) that can prune provably infeasible values from the variables in $S(c_k)$. A propagator for $c_k$ can be seen as a function:
\begin{equation}
\pi: (D_j \ | \ x_j \in S(c_k)) \mapsto (D^\prime_j \ | \ x_j \in S(c_k))
\label{eqn:propagator}
\end{equation}

\noindent where $(D_j \ | \ x_j \in S(C_k))$ is a n-ple with the domains of the variables in $S(c_k)$. For all of them it must hold $D^\prime_j \subseteq D_j$. Informally, the propagator maps a group of domains to a shrinked version of themselves. A CSP is typically solved via branching, by posting additional constraints and triggering their propagators. This causes a domain reduction, potentially awakening other propagators until a fix-point is reached. Optimization can be performed by adding, whenever a feasible solution is found, a permanent constraint that requires the future solutions to have a better cost.

\emph{Resource Constrained Project Scheduling Problems} (RCPSP) consist in finding a start time for a set $A$ of activities. Each activity $a_i$ has a fixed duration $d_i$ and requires an amount $r_{ik}$ of each resource $r_k$ from a set $R$. Each resource has limited capacity $cap_k$ and the activities may be connected by precedence constraints. The goal is to minimize the worst case completion time (makespan). A RCPSP is modeled in CP by introducing a start variable $s_i$ for each activity and by modeling the resource restrictions via \texttt{cumulative} constraints \cite{aggoun1993extending,baptiste2000constraint}, that enforce the following relation for each resource $r_k$: 
\begin{equation}
\sum_{s_i \leq t < s_i + d_i} r_{ik} \leq cap_k \quad\quad \forall t = 0..eoh
\end{equation}

\noindent i.e. no resource overusage can occur. The term $eoh$ refers to the maximum possible end time, where each end time $e_i$ corresponds to $s_i + d_i$. The bounds for $s_i$ and $e_i$ have conventional names: $est_i$ and $lst_i$ are the earliest and latest start times, respectively, while $ect_i$ and $lct_i$ are the earliest and latest completion (end) times. The \texttt{cumulative} constraint is one of the most studied in CP and has several propagators \cite{baptiste2001constraint,vilim2009edge,vilim2011timetable,ouellet2013time,Laborie2003}. In this document, we focus on Energetic Reasoning and (secondarily) on Timetable propagation.

\emph{Timetable propagators} base their deductions on compulsory parts, i.e. time intervals where tasks must necessarily be processed. By aggregating the compulsory parts of all activities we can obtain a minimum resource consumption profile. Based on this information, we can prune the domain of $s_i$ if we realize that scheduling the activity at the Earliest Start Time $\underline{s}_i$ would exceed the available capacity in the minimum consumption profile. Several Timetabling based algorithms exist, with complexity of $O(n^2)$ \cite{letort2012scalable,beldiceanu2002new}.


\emph{Energetic Reasoning} is a propagator based on the concept of energy consumption on a given time interval. If the minimum consumption is larger than the provided energy on this interval, the constraint cannot be satisfied or at least some bound adjustments can be done. The algorithm runs in $O(n^3)$. Note that there exist $O(n^2)$ algorithms that cannot perform bound adjustments and only detect inconsistencies \cite{baptiste2001constraint,derrien2013energetic}. Nevertheless, using the classical ER algorithm can still considerably reduce the search space compared to the combination of TT and the ER checker: on the BL instances, a reduction by a factor of 3 is achieved in $\sim 45\%$ of the instances with dichotomic search and on $\sim 19\%$ with \emph{SetTimes}.



\subsection{Algorithm Selection and Propagation} 
\label{sub:algorithm_selection_and_propagation}

The problem we consider is strictly related to Algorithm Selection, meaning the activity of deciding the best algorithm for tackling a given problem, that was first formalized in \cite{rice1975algorithm}. Since then, the field has received a lot of attention from the Optimization and ML communities and as a consequence the related literature is vast and complex.  There are off-line approaches, relying on problem features to choose either a single algorithm, or a set of algorithms to be executed in parallel (or in sequence according to a schedule). On-line approaches can adjust the selection at run-time, but incur additional overhead problems. The selection activity may involve picking altogether different algorithms or adjusting the parameters of a single approach. Different ML techniques have been employed, ranging from simple heuristic rules to statistical regression and more complex techniques such as Decision Trees, Artificial Neural Networks, Support Vector Machine and Clustering. For an excellent overview on Algorithm Selection in the context of Combinatorial Optimization (covering Hydra -- and derivatives --, SATzilla, ParamILS and ISAC), the reader is referred to \cite{kotthoff2012algorithm}. For selecting the best learning algorithm for a learning problem (so-called meta-learning) a nice overview is provided in \cite{smith2008cross}.

Despite the extensiveness of the literature about Algorithm Selection, only a few works so far have addressed the problem of choosing propagators (or adjusting the consistency levels) in CP automatically. The earliest example is \cite{el1996instance}, where the authors propose a method to detect when using simple Forward checking can achieve the same consistency level as Arc Consistency. The idea is generalized in the approach from \cite{borrett1996adaptive}, that switches back and forth between simpler (and faster) consistency algorithms and more powerful (and slower) ones depending on their observed and predicted performance. More recently, heuristic rules to switch between ``strong'' and ``weak'' propagation have been proposed in \cite{stergiou2009heuristics,balafrej2013adaptive}. The particular case of ER is considered in \cite{berthold2011approximative}, where an approximative criterion is used to estimate the potential of ER.

The use of Machine Learning methods for selecting propagators has been considered in \cite{gent2010machine}. In the paper, the authors use classification techniques to select which propagator (and which implementation) to use for the \texttt{alldiff} constraints on a given instance. The classifier is trained on a set of benchmark problems and takes as input general attributes of the instance (e.g. the number of \texttt{alldiff} constraints) and more complex features obtained from its primal graph. 





\section{Design Process} 
\label{sec:design_process}

\subsection{Problem Definition} 
\label{sub:problem_definition}


Given a target constraint $c_k$ and the current domains of the variables in its scope $S(c_k)$, we consider the problem of predicting whether a propagator $\pi$ will cause some pruning or not, with reasonable probability. Formally, we are interested in designing an oracle function $O_\pi$ such that:
\begin{equation*}
O_\pi(D_i | x_i \in S(C_k)) = \left\{\begin{aligned}
&true \text{ if some value is pruned}\\
&false \text{ otherwise} \end{aligned}\right.
\end{equation*}

\noindent One can see that $O_\pi$ has the same input as $\pi$ -- see Equation~\eqref{eqn:propagator}. The $O_\pi$ function is meant to be used as a guard condition for the execution of the propagator.

This problem formulation has a number of advantages: first, the oracle is \emph{guaranteed to have enough information to make a correct guess}. The challenge is therefore to devise an $O_\pi$ function with lower complexity than the propagator itself. Second, if a new propagator for the constraint is introduced, a new oracle must be trained, but \emph{the existing ones require no modification at all}. Third, the oracle \emph{can be checked at any point during search}, making the designer completely free about how to combine propagators (as long as the fallibility of the oracle is taken into account). More importantly, this also makes the approach well suited for use in complex search strategies and Large Neighborhood Search.

The simplest combination scheme for a set of propagator consists in running a lightweight algorithm (or even just a checker) until the fix-point is reached, and then running a single iteration of a more complex propagator only if $O_\pi$ returns true. This simple idea is similar to the one used in \cite{borrett1996adaptive}. As already mentioned, in this paper we consider the specific case of building an oracle function for ER, to be consulted once TT propagation has reached a fix-point.

In general, we propose to use ML techniques to obtain the oracle $O_\pi$ for complex propagators. In this context, obtaining the function requires to: 1) build a representative training set; 2) selecting features (based on the variable domains and on static information) for the classifier; 3) choosing a classification technique, then training and evaluating a ML model.

In the following, we will discuss those three steps in detail, with a focus on investigating the feasibility of an \emph{accurate} oracle function. Before proceeding, it is worth to note that it should be possible to improve the efficacy of our approach by accounting for temporal aspects (e.g. the propagation time) in the oracle definition. This is left for future research and discussed in the concluding Section~\ref{sec:conclusion_and_future_work}.

\subsection{Build a Representative Training Set} 
\label{sub:build_a_representative_training_set}


As a consequence of our problem definition, a (raw) training set consists of a number of tuples $(D_i \ |\ x_i \in S(c_k))$. In other words each tuple corresponds to a possible state for the domains of the constraint variables (the start times, for the ER propagator). Additionally, any static information concerning the constraint should be considered (e.g. the durations and the resource capacity). Finally, each tuple must be associated to the corresponding exact value of the $O_\pi$ function. 

A good training set should be representative of any realistic domain configuration. Such a set can be obtained by sampling the domains when solving a set of benchmarks problems, provided that:
\begin{itemize}
\item The considered benchmarks and the search strategy used for their solution are representative enough.
\item The samples are collected at representative times.
\item The samples come from diverse regions of the search tree.
\item The samples come from diverse constraints (e.g. with different scopes and different static information).
\end{itemize}

\noindent The exact meaning of the word ``representative'' depends on the context. In particular, we can distinguish between two application scenarios. First, one may want to obtain an $O_\pi$ function as general as possible (e.g. for inclusion in a constraint solver). In this case, the samples should be collected using many benchmarks and different search strategies. Alternatively, one may be interested in a particular benchmark, or a particular search strategy, or a particular combination scheme for the propagators. In this case, the samples should be collected only for the relevant cases, to increase the chance of obtaining an accurate oracle.

In this paper we take a mixed approach. Since we plan to use the $O_\pi$ function to decide whether to run ER after TT, we collect samples only after the latter has reached the fix-point. We also commit to a single branching strategy (\emph{SetTimes}), to avoid overcomplicating the learning problem (this is a feasibility study, after all) and because such search strategy is widely used in scheduling. We do however investigate the impact of using a single or multiple benchmarks for the training set (see Section~\ref{sec:xp}).

We consider scheduling problems with different number of activities, resources and different resource requirements. To avoid having overly similar tuples, at search time we perform the sampling with a small, fixed probability. This method decreases the risk of building a training set with many samples coming from nodes too close to each other in the search tree.

For quickly collecting samples on diverse regions of the search, we use Randomized Large Neighborhood Search \cite{godard2005randomized}. This may bias the oracle performance, but allows to experiment more easily with different training sets. Note that using Large Neighborhood Search incurs the risk of generating several times equivalent domains. It may be a good idea to remove such duplicates to ensure fairness in the training and in the evaluation phase.

\subsection{Input Features} 
\label{sub:input_features}

\newcommand{\relEHistogram}{\widetilde{E}(t)}
\newcommand{\CPEHistogram}{\widetilde{E}_{\mathit{CP}}(t)}

Selecting effective features for a classifier is ``one of the most important, yet nebulous, aspects of the algorithm selection problem'' \cite{rice1975algorithm}. In this section, we try to provide guidelines for picking meaningful features by describing the one we used for the oracle function for ER. In particular, we obtain our features by first extracting intermediate characterizations for the cumulative constraint (\textit{cumulative characterizations}), and then by computing aggregated statistics. The cumulative characterizations are numbers obtained from static information about the cumulative constraint and from the domains. For their description it is useful to introduce some definitions, introduced in \cite{baptiste2001constraint} and \cite{berthold2011approximative}:

\begin{itemize}
\item Point of interests considered by ER:
\begin{align*}
O_1 = \{ \est_i\} \cup \{\ect_i  \} \cup \{\lst_i \}\\
O_2 = \{ \lct_i\} \cup \{\lst_i\} \cup \{\ect_i \}
\end{align*}

\item Time Intervals considered by ER:
\begin{equation*}
\begin{array}{c}
\{ [t,t^\prime) \} \, \forall t \in O_1, \forall t^\prime \in O_2, t^\prime \geq t \\
\cup \,\{ [t,t^\prime) \} \, \forall t \in O_1, \forall t^\prime \in O(t), t^\prime \geq t \\
\cup \, \{[t,t^\prime) \} \, \forall t^\prime \in O_2, \forall t \in O(t^\prime), t^\prime \geq t 
\end{array}
\end{equation*}
where $O(t) = \{ \est_i + \lct_i - t\}$

\item  Relative Energy:
\[\tilde{E}_i = \frac{d_i \cdot r_{ik}}{lct_i - \est_i}\]

\item Relative Energy Histogram: 
\[ \relEHistogram = \sum_{a_i \in A : \est_i \leq t < \lct_i } \frac{d_i \cdot r_{ik}}{\lct_i - \est_i}\]
\end{itemize}

\noindent\textbf{Generic cumulative characterizations:} Table \ref{table:act_features} presents our generic cumulative characterizations. They are computed for each activity $a_i$ and have $O(1)$ complexity, so that for a given cumulative constraint we have a vector of values for each characterization type. Obtaining each vector has complexity $O(n)$, where $n$ is the number of activities.  The rationale for using \textit{normalizedCompulsoryPart} and \textit{domainTightness} is the fact that we run ER after TT and the intuition that bound adjustments from ER are more likely to happen when the domains of the start variables are small \cite{berthold2011approximative}.
 
 \begin{table}[ht]
 \centering
 \scalebox{0.8}{
 \begin{tabular}{|l|c|}
 \hline
 \textbf{Characterization Name} & \textbf{Value} \\ 
 \hline \hline
 normalizedCompulsoryPart  & $\frac{\mathit{max}(0,\est_i + \dur_i - \lst_i)}{(\lst_i + \dur_i - \est_i) }$ \\
 \hline 
 fixedActivity  & $\lfloor \est_i / \lst_i \rfloor $ \\
 \hline
 domainTightness  & $\mathit{max}(0, \lst_i - \est_i )$ \\
 \hline
 normalizedRelativeEnergy  & $\tilde{E}_i / C $ \\
 \hline
 est  & $est_i $ \\
 \hline
 lct  & $lct_i $ \\
 \hline
 duration  & $d_i $ \\
 \hline
 requirement  & $r_{ik} $ \\
 \hline
 actSlack  & $1 - r_{ik}/cap_k $ \\
 \hline
 \end{tabular}
 }
 \caption{Generic cumulative characterizations.} \label{table:act_features}
 \end{table}

\smallskip
\noindent\textbf{ER specific characterizations:} The characterizations in Table \ref{table:tp_features} are specific to ER. Both \textit{timePoints} and \textit{intervalSize} are computed for every Time Interval considered in ER. The notation $\mathit{lub}$ stands for the \textit{least upper bound} and is used instead of $\mathit{max}$ as the intervals are right-open. The characterization \textit{relativeEnergyTimePoints} is computed for every time point in the \textit{timePoints} characterization.
The size of \textit{timePoints} and \textit{relativeEnergyTimePoints} are $O(n)$ while the one of \textit{intervalSize} is $O(n^2)$. For each characterization, the time complexity for computing all the values is $O(n^2)$.

\begin{table}[ht]
\centering
\scalebox{0.8}{
\begin{tabular}{|l|c|c|}
\hline
\textbf{Characterization Name} & \textbf{Parameter} & \textbf{Value} \\
\hline \hline
timePoints & Interval $\cal{I}$ & $\{\mathit{min}(\cal{I}), \mathit{lub}(\cal{I})\}$ \\
\hline 
intervalSize & Interval $\cal{I}$ & $\mathit{lub}(\cal{I}) - \mathit{min}(\cal{I})$ \\
\hline
relativeEnergyTimePoints & Time point $t$ & $ \relEHistogram / C$ \\
\hline
\end{tabular}
}
\caption{ER specific characterizations.} \label{table:tp_features}

\end{table}

\smallskip
\noindent\textbf{Final features:} The features used as input for the classification algorithm are aggregation statistics computed for each characterization type. In particular, for each vector of values we consider the \textit{minimum}, \textit{maximum}, \textit{arithmetic mean}, \textit{geometric mean}, \textit{median}, \textit{first quartile}, \textit{third quartile}, \textit{population variance}, \textit{sample variance}, \textit{kurtosis}, \textit{skewness}, \textit{length}, \textit{cardinality} (i.e., the number of distinct elements).


\smallskip
\noindent\textbf{Complexity:} The time complexity to compute a cumulative characterization is at most $O(n^2)$, lower than the one of ER. Most of the statistics that we employ as features have complexity $O(m)$, where $m$ is the size of the vector for which the feature is extracted. Some statistic operations involve an ordering step and have therefore a complexity of $O(m\log(m))$. Hence, the worst time complexity for the feature computation is $O(m^2.log(m^2))$, but most of them are actually obtained in $O(n)$, $O(n\log(n))$ or $O(n^2)$.


\subsection{Selecting and Evaluating a Classifier} 
\label{sub:selecting_and_evaluating_a_classifier}

Any classification technique can in principle be used to obtain an oracle function. In this work (a feasibility study) we focus on Random Forests and Support Vector Machines for their prediction power. In a more practical setting, the best classifiers will be those with a good trade-off between prediction power and computation cost of the classification. We evaluate the classifiers on a portion of the set from Section~\ref{sub:build_a_representative_training_set}, that is not employed for the training phase. We measure the performance based on $TP$, $TN$, $FP$, $FN$ representing the number of tuples that are correctly classified as $true$, correctly classified as $false$, incorrectly classified as $true$, incorrectly classified as $false$. Then we consider the following metrics:

\begin{itemize}
\item The $\mathit{Fallout} = \FP / (\FP + \TN)$, corresponding to the number of erroneous activations of ER over all the cases when it is not supposed to run. The lower the metric, the fewer times were are paying the $O(n^3)$ cost of ER in vain.
\item The $\mathit{Recall} = \TP / (\TP + \FN)$, corresponding to the number of times ER is triggered when it is indeed supposed to run. The larger the metric, the more benefits we have from the propagation power of ER.
\end{itemize}

\section{Experiments}
\label{sec:xp}

\newcommand{\dsbl}{\mathit{ds}_\mathit{bl}}
\newcommand{\dspack}{\mathit{ds}_\mathit{pack}}
\newcommand{\dsmerge}{\mathit{ds}_\mathit{merge}}





\subsection{Training Set and Classifiers} 
\label{sub:training_set_and_classifiers}

We obtained our training set as described in Section~\ref{sub:build_a_representative_training_set}, by solving via LNS the scheduling problems in the BL benchmark \cite{baptiste2000constraint} and the 20 first elements of the Pack benchmark \cite{artigues2007resource}. The considered instances differ for the number of activities (15 to 33), the resource capacities and requirements. For each instance, we performed 15 LNS iterations, each stopped after 500 fails. At each iteration, roughly 20\% of the activities were relaxed. For each search node, we sampled the domain state of all constraints with a 0.1 probability. Duplicates are removed to prevent unfair results. This resulted in the generation of a data set of 6515 elements with 20.87\% of cases with $O_\pi = true$ for the BL benchmark, and a data set of 13979 elements with  20.50\% of cases with $O_\pi = true$ for the Pack instances. We refer to those data set as $\dsbl$ and $\dspack$. We also built a data set by merging $\dsbl$ and $\dspack$ and called it $\dsmerge$.



Since in this work we are primarily interested in the accuracy of $O_\pi$, we decided to focus on Random Forests -- RF, \cite{breiman2001random} -- and Support Vector Machines -- SVM, \cite{cortes1995support} -- because of their prediction capabilities when the set of most relevant features is not clearly known. In particular, we used the \textit{randomForest} \cite{randomForest} and \textit{e1071} \cite{e1071} R packages.

Random Forests generally build CART trees \cite{breiman1984classification} with no pruning. The complexity for classifying a new instance (which is part of the overhead of our approaches) depends on the actual trees that are learned. A rather pessimistic approximation of the complexity is $O(\mu\log(\nu))$, where $\mu$ is the number of trees and $\nu$ is the number of features. 

A trained linear SVM has a time complexity linear in the number of features (scalar product evaluation).
For a non-linear kernel function, the classification complexity is linear in the number of support vectors (SV's). The complexity of a SVM model thus scales with the most difficult samples, forcing an increase in Support Vectors\footnote{In \cite{fehr2008fast}, they speed up the classification by reducing the number of SV's.}.






\subsection{Evaluation} 
\label{sub:evaluation}


We trained (using the default parameters) and tested classifiers using the data sets $\dsbl$ and $\dspack$ (25\% of the set was used for the evaluation). In Table~\ref{table:conf_inter:bl} and \ref{table:conf_inter:pack} we report the 95\% confidence intervals for the Fallout and the Recall metrics. All the tested classifiers (in particular the Random Forest) have good performance, with low Fallout and large Recall values.

\begin{table}[ht]
\centering
\caption{ 95\% confidence intervals for Fallout and Recall on the two benchmarks.} \label{table:conf_inter}
\scalebox{0.75}{
\begin{tabular}{|l||c|c|c|}
\hline
Classifier & Fallout & Recall \\
\hline
Random Forests & $[0.0076 , 0.0204]$ & $[0.8918 , 0.9493 ]$\\ 
\hline
SVM (gaussian) & $[0.0276 , 0.0485 ]$& $[0.7385 , 0.8262 ]$\\ 
\hline
SVM (linear) & $[0.0364, 0.0598 ]$  & $[0.7607, 0.8452 ]$\\ 
\hline
\end{tabular}
}
\subcaption{$\dsbl$} \label{table:conf_inter:bl}


\scalebox{0.75}{
\begin{tabular}{|l||c|c|c|}
\hline
Classifier & Fallout & Recall \\
\hline
Random Forests & $[0.0103 , 0.0192] $& $[0.9484 , 0.9763 ]$\\ 
\hline
SVM (gaussian &  $[0.0242 , 0.0370] $& $[0.9101 , 0.9477]$ \\ 
\hline
SVM (linear) & $ [ 0.0245 , 0.0374 ]$ & $[0.9258 , 0.9598 ]$ \\ 
\hline\end{tabular}
}
\subcaption{$\dspack$}\label{table:conf_inter:pack}
\end{table}

Then, in order to put to the test the generalization ability of the classifiers we tried a cross-benchmarks experiments. Namely, we evaluated the classifiers trained on $\dsbl$ dataset over the $\dspack$ one, and vice-versa. The results were not satisfactory. This could be a consequence of the training set not being representative enough, or it could hint at the presence of many non meaningful features in the classifier input.

With the aim to investigate the possible causes, we tried to use the combined $\dsmerge$ set for the training, and then we tested the obtained classifiers on (previously unseen) instances from both $\dsbl$ and $\dspack$ (25\% in both cases). The confidence intervals of Fallout and Recall for this experiment are given in Table~\ref{table:conf_inter:merge:bl} and \ref{table:conf_inter:merge:pack}. As one can see, the performance of the classifiers are now only slightly worse than the original ones. This suggests that designing a representative training set (in particular, including diverse benchmarks) has a strong impact on the classifier ability to generalize.




\begin{table}[ht]
\centering
\caption{ 95\% confidence intervals for Fallout and Recall on the two benchmarks, for classifiers trained over $\dsmerge$.}

\scalebox{0.75}{
\begin{tabular}{|l||c|c|c|}
\hline
Learning algorithm & Fallout & Recall \\
\hline
Random Forests & $[0.0113 , 0.0261]$ & $[0.9067 , 0.9595 ]$\\ 
\hline
SVM (gaussian) & $[0.0236 , 0.0433 ]$& $[0.7760 , 0.8577 ]$\\ 
\hline
SVM (linear) & $[0.0476, 0.0738 ]$  & $[0.7012, 0.7930 ]$\\ 
\hline \end{tabular}
}
\subcaption{$\dsmerge$ tested on a disjoint part of $\dsbl$} \label{table:conf_inter:merge:bl}

\scalebox{0.8}{
\begin{tabular}{|l||c|c|c|}
\hline
Learning algorithm & Fallout & Recall \\
\hline
Random Forests & $[0.0091 , 0.0178]$ & $[0.9472 , 0.9750 ]$\\ 
\hline
SVM (gaussian) & $[0.0254 , 0.0386 ]$& $[0.9059 , 0.9438 ]$\\ 
\hline
SVM (linear) & $[0.0300, 0.0442 ]$  & $[0.75910, 0.9318 ]$\\ 
\hline\end{tabular}
}
\subcaption{$\dsmerge$ tested on a disjoint part of $\dspack$}\label{table:conf_inter:merge:pack}
\end{table}

\begin{figure}[bt]
\centering
\includegraphics[scale=0.2]{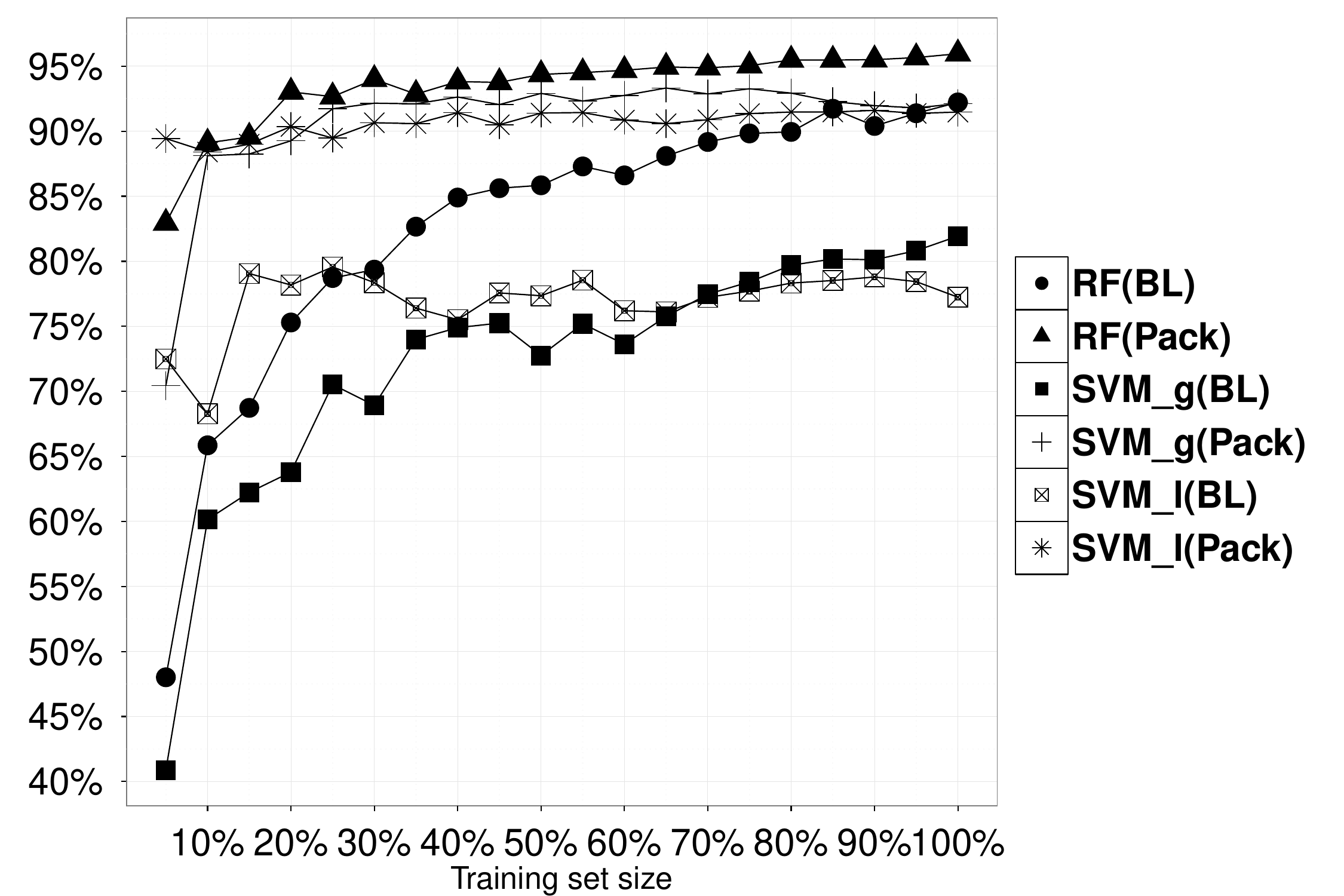}
\caption{Average Recalls computed on cumulative training sets constructed from $\dsbl$ and $\dspack$}\label{img:recalls}
\end{figure}

\begin{figure}[tb]
\centering
\includegraphics[scale=0.2]{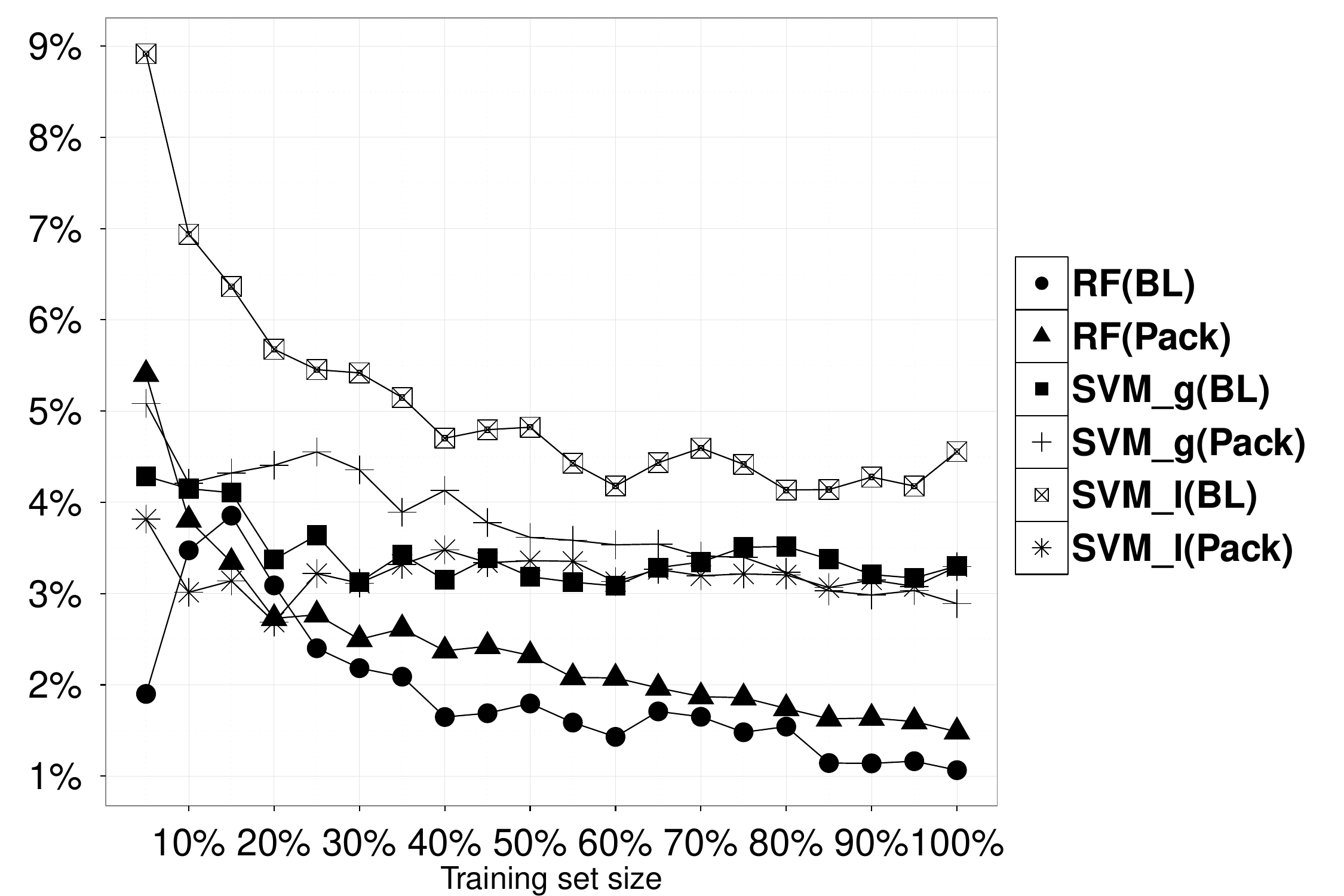}
\caption{Average Fallouts computed on cumulative training sets constructed from $\dsbl$ and $\dspack$}\label{img:fallouts}
\end{figure}

\noindent\textbf{Warm-up:} The slightly decreased performance observed in Table~\ref{table:conf_inter:merge:bl} and \ref{table:conf_inter:merge:pack} suggests also that, in case the focus is on obtaining the best results for a specific benchmark, an excessive diversity of the training set may have adverse effects. In this setting, it would also be particularly interesting if the classifiers could keep a good predictive power, even when trained on a small dataset. This would allow to tackle a scheduling benchmark by running a short ``warm-up'' phase, for training a classifier to be used when solving the remaining instances. We tested the feasibility of this approach on the two sets $\dsbl$ and  $\dspack$ separately. We divided those data sets in several folds to create training sets of increasing sizes. Each of them was used to compute average metrics using 10-fold cross-validation. Figure~\ref{img:recalls} and \ref{img:fallouts} show respectively the average Recall and Fallout as a function of the training set size for several approaches. As one can see, reasonably good prediction rates are obtained with Random Forests when the training set contains as little as 25\% of the total number of tuples.

\noindent\textbf{Feature importance:} Proper feature selection is a complicated task and out of the scope of this paper. However, the \textit{importance} scores \cite{breiman2001random} given to attributes in Random Forests provides an estimate of how discriminant the features are. In a final experiment, we considered for $\dsmerge$ the best features (15 out of 137) , according to their reported Gini index and average accuracy. We removed from such list the feature \emph{Median(relativeEnergyTimePoints)}, because of its large computational complexity, and then we tried to learn classifiers on $\dsmerge$ based only on such features. Table \ref{table:best_f_predict} reports the 95\% confidence intervals of Fallout and Recall for the experiment. As one can see, a small subset of the features is sufficient to reach good prediction accuracy (in particular with Random Forests). This enables a considerable reduction of the overhead when the oracle function is embedded in a constraint solver. In the considered case, the retained features are far fewer than the original ones and their maximum complexity is $O(n^2)$. Incremental computation could also be used to amortize the computational complexity.


%

\begin{table}[ht]
\centering
\scalebox{0.75}{
\begin{tabular}{|l||c|c|c|}
\hline
Learning algorithm & Fallout & Recall \\
\hline
Random Forests & $[0.0190 , 0.0294]$ & $[0.8802 , 0.9219 ]$\\ 
\hline
SVM (gaussian) & $[0.0349 , 0.0486 ]$& $[0.7476 , 0.8057 ]$\\ 
\hline
SVM (linear) & $[0.0549, 0.0715 ]$  & $[0.6463, 0.7115 ]$\\ 
\hline
  \end{tabular}
}
\caption{95 \% confidence intervals of Fallout and Recall metrics of classifiers built using only the best features.} \label{table:best_f_predict}
\end{table}

\section{Conclusion and Future Work} 
\label{sec:conclusion_and_future_work}

This paper proposes an approach to take better advantage of complex propagators, by running them only when they provide an actual benefit in terms of pruning. The main idea is that of \emph{paying for what you get}, so as to make the best possible use of powerful propagators often forsaken for their low scalability. We propose to achieve this goal by relying on an oracle function, obtained via ML techniques. A combination of design decisions makes our approach particularly flexible compared to existing alternatives. In this work we focus on investigating the feasibility of an accurate oracle for ER in the context of the \texttt{cumulative} constraint. We show that a very high prediction accuracy can be obtained a reasonably low time complexity.

Future work involves in first place making the approach faster, by relying on fewer or cheaper-to-compute features, by using incremental computation, or by simplifying the classifiers (e.g. fewer trees in a Random Forest). Reducing the training time and the size of the required training set are also important factors when using a ``warm-up'' approach before tackling a large set of instances. Second, we obviously plan to embed the oracle function in an actual propagator and test its efficacy when solving a variety of scheduling problems. In this context, we also plan to improve the efficacy of the oracle function by taking into account the trade-off between the propagation complexity and the expected benefit of pruning values \emph{in terms of solution time}. Finally, our approach should be tested on other constraints and propagators, and compared with alternative techniques (e.g. state base heuristics). A investigation of alternative ML based configurations may also be interesting. For example, using a classifier to directly pick a propagator, rather than to predict its pruning power, may incur less overhead (at the price of reduced flexibility).

\bibliographystyle{aaai}
\bibliography{biblio}
\end{document}